\documentclass[letterpaper, 10 pt, journal, twoside]{ieeetran}
\IEEEoverridecommandlockouts
\usepackage{geometry}
\usepackage{cite}
\usepackage{amsmath,amssymb,amsfonts, dsfont}
\usepackage{algorithmic}
\usepackage{graphicx}
\usepackage{textcomp}
\usepackage{xcolor}
\usepackage{bbm}
\usepackage{multirow}
\usepackage[linesnumbered,ruled,vlined]{algorithm2e}
\usepackage{soul}
\usepackage{svg}
\usepackage{array}
\svgpath{{fig/assets/}} 
\sethlcolor{yellow}
\def\BibTeX{{\rm B\kern-.05em{\sc i\kern-.025em b}\kern-.08em
    T\kern-.1667em\lower.7ex\hbox{E}\kern-.125emX}}
\DeclareMathOperator*{\argmax}{arg\,max}
\begin{document}

\setlength\extrarowheight{1pt}
\setlength{\tabcolsep}{3pt}

\title{Constraining Gaussian Process Implicit Surfaces for Robot Manipulation via Dataset Refinement\\
\thanks{
Manuscript received: May 28 2024; Revised: Aug. 8 2024; Accepted: Sept 12 2024.
This paper was recommended for publication by Editor Júlia Borràs Sol upon evaluation of the Associate Editor and Reviewers’ comments. 
$^1$ Robotics Department, University of Michigan, Ann Arbor, USA \{abhin, pmitrano, dmitryb\}@umich.edu.This work was supported in part by the Office of Naval Research Grant N00014-24-1-2036 and NSF grants IIS-2113401 and IIS-2220876.
Digital Object Identifier (DOI): see top of this page.}
}
\markboth{IEEE Robotics and Automation Letters. Preprint Version. Accepted September, 2024}
{Kumar \MakeLowercase{\textit{et al.}}: Constraining GPIS via Dataset Refinement}

\author{Abhinav Kumar$^1$, Peter Mitrano$^1$, Dmitry Berenson$^1$}

\maketitle

\begin{abstract}
Model-based control faces fundamental challenges in partially-observable environments due to unmodeled obstacles.
We propose an online learning and optimization method to identify and avoid unobserved obstacles online.
Our method, Constraint Obeying Gaussian Implicit Surfaces (COGIS), infers contact data using a combination of visual input and state tracking, informed by predictions from a nominal dynamics model.
We then fit a Gaussian process implicit surface (GPIS) to these data and refine the dataset through a novel method of enforcing constraints on the estimated surface.
This allows us to design a Model Predictive Control (MPC) method that leverages the obstacle estimate to complete multiple manipulation tasks.
By modeling the environment instead of attempting to directly adapt the dynamics, our method succeeds at both low-dimensional peg-in-hole tasks and high-dimensional deformable object manipulation tasks.
Our method succeeds in 10/10 trials vs 1/10 for a baseline on a real-world cable manipulation task under partial observability of the environment.
\end{abstract}

\vspace{-.7cm}
\section{Introduction}
\vspace{-.2cm}
\IEEEPARstart{S}{pecial} care must be taken when using model-based planning and control methods in partially observable environments.
This is particularly important where not all obstacles are modeled by dynamics, to avoid collisions with unmodeled or unobserved parts of the environment.
Such collisions could prevent task completion; for instance, the object being manipulated might be blocked by the unmodeled environment object.
The challenge is heightened when manipulating deformable objects like cables in the home or office.
These objects can interact with unmodeled parts of the environment in complex ways due to high-dimensional, highly nonlinear dynamics.
This creates more possibilities for task failure.

Prior work has explored ways to model objects in the environment based on data from partial visual observations and/or contact \cite{CLASP, smith20203d, smith2021active, suresh2022shapemap}.
However, such estimates can produce inaccuracies that may lead to the task becoming infeasible (e.g. blocking the path to the goal).
In this work, we introduce Constraint Obeying Gaussian Implicit Surfaces (COGIS).
COGIS uses a Gaussian process implicit surface (GPIS) \cite{GPIS} to model obstacles using contacts inferred during task execution.
It also uses a novel optimization approach to ensure the obstacle surface satisfies provided constraints.

\begin{figure}[t]
\centerline{\includesvg[inkscapelatex=false, width=\linewidth]{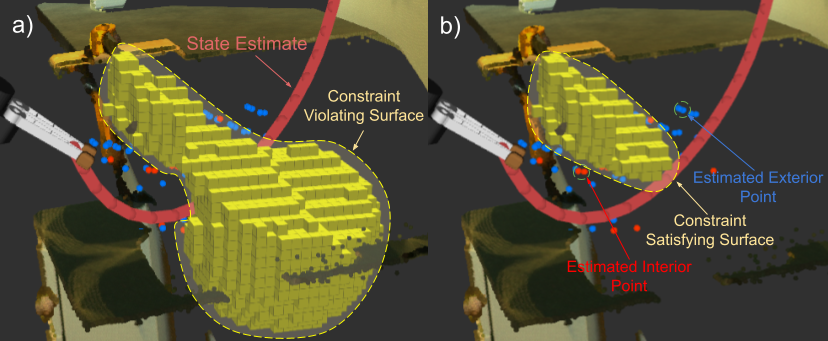}}

\caption{Our method learns a continuous model of the obstacle geometry as an implicit surface, voxelized here for visualization, while enforcing constraints on the model. We model contacts as pairs of points interior and exterior to the 0-level-set surface. \textbf{a)} A constraint violating surface where the cable state estimate penetrates the surface due to noisy estimates of interior and exterior points. \textbf{b)} The surface after estimated contacts have been refined.}
\label{fig:grabber}
\end{figure}

COGIS learns a GPIS using contacts estimated by tracking the state of a manipulated object.
It also incorporates predictions from nominal dynamics and visual data.
This enables obstacle modeling without specialized tactile sensing, which may not be available along the surface of a manipulated object.

The GPIS in COGIS uses a Gaussian Process (GP) to learn a 0-level-set surface that we use to model obstacle geometry.
The underlying GP provides a method of estimating surfaces from the estimated contacts along with an uncertainty estimate. Its kernel function also provides a useful inductive bias that encourages smooth surface predictions.

By optimizing the contact dataset, we enforce user-provided constraints on the GPIS without assumptions on the form, convexity, or differentiability of the constraints.
We do this with CMA-ES with Margin \cite{cma}, a particle-based optimizer.
These constraints can incorporate domain knowledge or assumptions related to the task being performed.
Our key insight is that this task-specific information can be used to constrain estimated environment models, thereby improving task performance.
We use the estimated surface to construct costs for MPC.
Considering the estimated surface in the cost function, along with considering visible obstacles in the dynamics, allows the controller to navigate through the environment.

The contributions of this paper are:
\vspace{-.05cm}
\begin{itemize}
    \item A method for estimating obstacle geometries online using a fusion of visual input and contacts inferred through state tracking and predictions from nominal dynamics
    \item A method for ensuring the estimated geometry satisfies arbitrary task-informed constraints
\end{itemize}
\vspace{-.05cm}

We show that our method is able to identify obstacles and enable task completion for low-dimensional peg-in-hole tasks and high-dimensional deformable object manipulation tasks.
Baselines that do not adapt online or that attempt to reason in the dynamics space succeed less frequently for the higher-dimensional tasks.
We also show that enforcing constraints on the learned surface improves task performance.

\vspace{-.4cm}
\section{Related Work}
\vspace{-.3cm}
\textbf{Online Adaptation} An alternative to estimating environment geometry when using a pre-existing dynamics model is to account for unanticipated contact by adapting the dynamics model \cite{deisenroth2011pilco, Focus, wang2022offline,van2020online, lagrassa2022learning}.
These works directly update a learned dynamics model with data from the online environment or learn a residual dynamics model to capture the novel dynamics. 
While these methods are useful for low-dimensional state-action spaces, they require either multiple trials to collect sufficient data when dynamics are high-dimensional or fit simple linear models online.
In contrast, we attempt to estimate unseen obstacles, which exist in the 3D workspace and thus do not require adapting high-dimensional dynamics models.
We also do not require large amounts of data and our results show that our method can estimate novel objects well enough to complete tasks in a single episode.

TAMPC \cite{TAMPC} is a method that adapts MPC techniques to novel environments through identifying and avoiding traps, or local minima, of controllers, which sometimes arise from unanticipated obstacles.
TAMPC defines local minima in state-action space.
However, in manipulation problems, local minima are often induced by geometric properties of the scene. 
By directly considering the geometry of obstacles and manipulated objects, we can learn a richer model that results in more efficient task execution.

\textbf{Contact Detection} There is prior work that estimates locations of contact points \cite{contact_particle_filter} or estimates properties of manipulated objects using contact \cite{9562060}.
These methods use estimates of joint torques and knowledge of robot geometries to calculate a belief over contacts.
These methods are not generally applied to deformable objects as we lack good torque estimates along an object like a cable. 
Lack of this data motivates alternate methods for contact estimation.

\textbf{Learned object/environment modeling} Learned implicit surfaces have previously been used to model environment and object geometries. 
Two popular classes of models are Gaussian process implicit surfaces (GPIS)\cite{8794324, gandler2020object, caccamo2016active, liu2021active, ottenhaus2019visuo, 7803372, dragiev2011gaussian, wu2023log, 9562060} and neural implicit surfaces (NIS)\cite{10160217, wi2022virdo, 9812146, 9976191}.
These works assume access to rich perception signals, including visual data with dynamic viewpoints or tactile sensors.
We do not assume that our viewpoint of the system can change over time nor do we have access to tactile data when grasped objects make contact with obstacles.
We instead rely on a combination of limited visual data with contacts estimated through state tracking and predictions from nominal dynamics.

\textbf{Constrained Implicit Surfaces}
Prior work has investigated methods for imposing constraints on shape reconstructions. 
One class of these methods can take the form of fitting parameterized functions to provided point clouds \cite{liu_constrained_2006, karniel_decomposing_2005, kovacs_constrained_2020}. These methods employ a fixed set of constraints and construct equations that allow them to satisfy these constraints within a convex optimization approach.
Alternate methods in rendering constrain implicit surfaces to respect haptic interaction \cite{haptic} and methods that learn 3D representations from 2D images include regularization terms to constrain certain geometric properties \cite{2D3D}.
In contrast to these approaches, our method imposes constraints on GPIS, handles arbitrary constraints, and fits the surfaces without access to multi-view visual data.

\vspace{-.3cm}
\section{Problem Statement}
\vspace{-.15cm}
In this paper, we consider manipulation problems in which a robot arm grasps an object and navigates it to a goal location.
We assume task execution begins with the object being grasped and the grasp is maintained throughout execution.

We consider the problem of optimal control in a partially observable environment.
Let $\textbf{u} \in \mathcal{U}$ represent the robot's action and $\textbf{X} \in \mathcal{X}$ represent the state of a grasped object.
We define $\textbf{X}=(\mathbf{x}^1,...,\mathbf{x}^n)$, meaning $\textbf{X}$ is represented as an ordered set of $n$ components where $\mathbf{x}^i \in \mathbb{R}^3$. This representation is useful for high-dimensional systems like deformable objects, which can be represented as a collection of particles or points of interest. For example, a cable can be represented as a set of ordered points in $\mathbb{R}^3$.
It can be applied to other systems where $n=1$, for example a peg grasped by a robot in a peg-in-hole task.
Using this representation enables independent reasoning about collisions between different components of manipulated objects and the environment.
We assume access to a function $d_\mathbf{x}(\textbf{x}_1, \textbf{x}_2)$ that provides a distance between state components. 

\begin{figure*}
\label{fig:block}
\centerline{\includesvg[inkscapelatex=false, width=\linewidth]{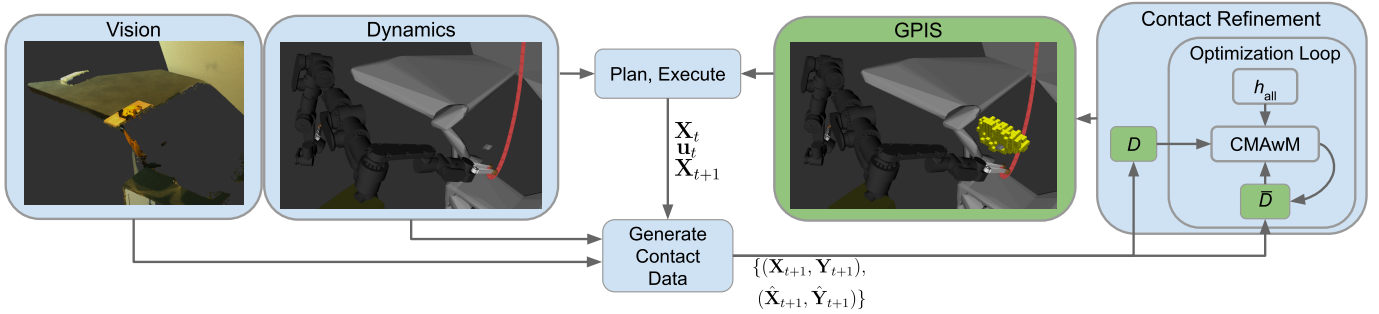}}
\vspace{-.35cm}
\caption{Block diagram showing the algorithm. Green blocks refer to objects generated by COGIS. The dynamics model, shown here as a MuJoCo simulation, and visual input are created offline and used to plan a trajectory along with the current GPIS. After an action is executed, we infer contacts from the transition $(\textbf{X}_t, \textbf{u}_t, \textbf{X}_{t+1})$, which we use to update the GPIS. The generated data $D$ are refined using CMAwM by selecting a subset $\bar{D}$ that ensures the GPIS satisfies provided constraints $h_{\mathrm{all}}$. We fit a surface in yellow that approximates the obstacle geometry occluded by the table while constraining the surface to avoid penetration with the cable state estimate.}
\vspace{-.7cm}
\end{figure*}

Given an initial state $\textbf{X}_0$ and a reachable goal set $G$, we seek a trajectory $\tau$ using (MPC) that reaches $G$ with a minimal number of control steps.
$G$ specifies known goal locations for a subset of components of $\textbf{X}$.
We refer to the goal location for a specific component $i$ as $G_i$.
This is useful when not all components have desired goal configurations, for example in a task where one end of a cable needs to be plugged in.

A trajectory $\tau$ has a horizon $T$, a sequence of controls $\tau_{\mathbf{u}}=\{\mathbf{u}_0 ... \mathbf{u}_{T-1}\}$, and a sequence of states $\tau_{\mathbf{X}}=\{\mathbf{X}_0 ... \mathbf{X}_T\}$.
A nominal dynamics model $f(\textbf{X}, \textbf{u})$ predicts the next state given the current state and some action.
$f$ is assumed to be provided and will be applied to a novel environment with obstacles that it may not be able to model due to partial observability.
We assume that error in $f$ is caused by unobserved obstacles.

Our method can utilize, but does not assume access to, a depth image $Z$ and corresponding point cloud $P$ of the environment collected without occlusion from the robot or manipulated object prior to task execution. 
Pre-generated depth images and point clouds prevent visibility issues caused by robot occlusion.
We use $d(\mathbf{X}, P)$ to refer to the minimum distance between each component of $\mathbf{X}$ and the points in $P$.

We assume the environment is static over the course of task execution.
We assume the robot and the grasped object can make contact with the environment without ceasing task execution or damaging itself or the environment.
This can be realized through compliant control or the ability to sense torques at joints of the robot and recover to a safe configuration when a threshold is crossed.

We assume the environment is partially observable and therefore partially modeled at initialization.
We seek to fit a model $\hat{\mathcal{E}}$ of the unknown environment geometry $\mathcal{E}$ while obeying one or more provided constraints $H = \{h_1, ..., h_n\}$.
$\hat{\mathcal{E}}$ will be fit from data collected during task execution.

This problem is challenging as estimating the model during task execution means we have access to limited data and no prior knowledge of the occluded region of the environment.
In addition, we do not assume tactile sensing is available. 

To address data quality issues, we incorporate explicit constraints into the estimation of $\hat{\mathcal{E}}$.
We define $h_{\mathrm{all}}= h_1 \land ... \land h_{|H|}$ as a constraint satisfied if all constraints in $H$ are satisfied.
These constraints can encode desired topological properties of the model, for example requiring there to be a collision-free path from the current state to $G$ given $\hat{\mathcal{E}}$. Our goal is to inform an MPC method using $\hat{\mathcal{E}}$. Rather than focusing on obtaining an accurate geometry of the unseen object, we only seek an $\hat{\mathcal{E}}$ that is sufficient for completing the task.

\vspace{-.2cm}
\section{Method}
\vspace{-.2cm}
Our method, shown in Fig. 2, can be split into two parts: Generating contact data to be used to estimate $\hat{\mathcal{E}}$ online (Secs. A, B, C) and refining $\hat{\mathcal{E}}$ to satisfy constraints (Sec. D).
We also include a description of our controller and how it uses $\hat{\mathcal{E}}$ to complete the task (Sec. E).
We execute these steps in a loop, shown in Alg. \ref{alg:cap}.
We first define the GPIS model used to fit $\hat{\mathcal{E}}$. A GPIS model learns a 0-level-set surface given exterior, surface, and interior points and their corresponding semantic labels:

\vspace{-.6cm}
\begin{equation} \label{GPIS_def}
  \texttt{GPIS}: \mathbb{R}^{3} \rightarrow \mathbb{R}; \texttt{GPIS}(\textbf{x})
    \begin{cases}
      < 0 & \text{if $\textbf{x}$ is interior}\\
      = 0 & \text{if $\textbf{x}$ is on the surface}\\
      > 0 & \text{if $\textbf{x}$ is exterior}
    \end{cases} 
\end{equation}

We define a novel optimization problem for the dataset refinement in the case of implicit surface models parameterized by a set $\bar{D}$ of points with corresponding semantic labels.
$\bar{D}$ is a subset of all collected data $D$.
As $\hat{\mathcal{E}}$ is parameterized by $\bar{D}$, updating $\bar{D}$ can be considered equivalent to updating $\hat{\mathcal{E}}$.

We seed the GPIS at initialization with the points in $G$ with corresponding labels of 1, reflecting our assumption that these points are reachable and therefore exterior to the surface.
The GPIS parameters are updated using gradient descent every $T_{\mathrm{fit}}$ iterations (see Table II).

\vspace{-.4cm}

\subsection{Dynamics-Informed Contact Data Generation}
We use predictions from nominal dynamics to estimate contacts which we then use to generate interior and exterior points of the GPIS.
By identifying regions of state space where the dynamics are inaccurate, we can generate candidate contact estimates.
While not all non-nominal dynamics are the result of contact, we can generate data in this manner and then refine contact estimates by enforcing constraints on the surface.

As we compute and execute control inputs using our controller discussed in Section \ref{sec:control}, we observe transitions of the form $(\textbf{X}_t, \textbf{u}_t, \textbf{X}_{t+1})$.
We estimate contacts by comparing the observed next state $\textbf{X}_{t+1}$ to the corresponding dynamics prediction $\hat{\textbf{X}}_{t+1}$ for the transition.
We generate labels $\textbf{Y}_{t+1},\hat{\textbf{Y}}_{t+1} \in \mathbb{R}^n$ for $\textbf{X}_{t+1}$ and $\hat{\textbf{X}}_{t+1}$ respectively, using Equations \eqref{eq:y'} and \eqref{eq:hat_y'}.

\vspace{-.4cm}
\begin{equation}
\label{eq:y'}
\mathbf{Y}_{t+1}^i = \mathrm{min}\left(\frac{d_\mathbf{x}(\mathbf{X}_t^i, \mathbf{X}_{t+1}^i)}{d_\mathbf{x}(\mathbf{X}_t^i, \mathbf{\hat{X}}_{t+1}^i)}, 1\right)
\end{equation}
\begin{equation}
\label{eq:hat_y'}
\mathbf{\hat{Y}}_{t+1}^i = 2\mathbf{Y}_{t+1}^i - 1
\end{equation}
\vspace{-.6cm}

\noindent As the values in $\mathbf{X}_{t+1}$ correspond to the tracked positions of the grasped object in the world, we can assume these points are exterior to or in contact with the surface.
Per \eqref{GPIS_def}, this means they should have a corresponding label in $\mathbf{Y}_{t+1} \geq 0$. 
\eqref{eq:y'} will compute a label between 0 and 1.
We interpret lower values as meaning a contact is more likely as motion is impeded.
An illustration of this can be seen in Fig.~\ref{fig:label_gen}a. We generate $2n$ potential data points for the GPIS corresponding to points in $\textbf{X}_{t+1}$ and $\mathbf{\hat{X}}_{t+1}$ per control step.
We add a subset of these points to the GPIS as explained in Section \ref{sec:add}.

A transition corresponds to a contact when $\mathbf{\hat{Y}}_{t+1}^i \leq 0$, which occurs when $\mathbf{Y}_{t+1}^i \leq .5$ per \eqref{eq:hat_y'}.
In this case, $\mathbf{\hat{X}}_{t+1}^i$ would correspond to an interior point.
Adding interior and exterior points with their corresponding labels to the GPIS allows it to interpolate a 0-level-set, thereby fitting a surface.

\begin{algorithm}[t]
\caption{High-Level Control Loop}\label{alg:cap}
Given $G, f, \hat{\mathcal{E}}, \texttt{MPC}, \alpha, \beta, \eta, C, f, d_{\mathbf{x}}, P, Z, r_c$\\
$\mathbf{X}_s$ = $\mathbf{X}_0$ \tcp{Saved state for local minimum detection}
\While{Not Reached Goal} {
    \If{$T_e$ steps since last component selection} {
        $s \gets \mathrm{argmin}\ \hat{\mathcal{E}}(\mathbf{X}_t)$\\
    }
    $\mathbf{u}_t \gets \texttt{MPC}(\mathbf{X}_t, G, \alpha, \beta, \eta, C, s, f, \hat{\mathcal{E}}, d_{\mathbf{X}})$\\
    $\mathbf{X}_{t+1} \gets $ apply $\mathbf{u}_t$, step environment\\

    $\mathbf{\hat{X}}_{t+1} \gets f(\mathbf{X}_t, \mathbf{u}_t)$\\
    Generate $\textbf{Y}_{t+1},\hat{\textbf{Y}}_{t+1}$ using Equations \eqref{eq:y'}, \eqref{eq:hat_y'}\\
    \If{$T_m$ steps since last local minimum check} {
        \texttt{local\_minimum} $\gets \frac{1}{T_m}\frac{1}{n}\sum_{i=1}^n d_\mathbf{x}(\mathbf{X}_{t+1}^i, \mathbf{X}_s^i) < d_{\mathrm{min}}$\\
        $\mathbf{X}_s \gets \mathbf{X}_{t+1}$
    } \Else{
        \texttt{local\_minimum} $\gets$ False
    }
    
    \texttt{pre\_process\_data}($\mathbf{X}_{t+1}, \textbf{Y}_{t+1},$ \\ 
    $~~~~\mathbf{\hat{X}}_{t+1}, \hat{\textbf{Y}}_{t+1}, P, Z, r_c, \texttt{local\_minimum}$)\\
    Update $D,~\bar{D}$ with $\{(\mathbf{X}_{t+1}, \textbf{Y}_{t+1}), (\mathbf{\hat{X}}_{t+1}, \hat{\textbf{Y}}_{t+1})\}$\\
    \If{$\neg h_{\mathrm{all}}$} {
        $\texttt{refine\_contacts}(D, \bar{D}, T_{\texttt{CMA}}, N, h_{\text{all}})$
    }
}
\end{algorithm}

\begin{algorithm}[t]

Given $\mathbf{X}_{t+1}, \textbf{Y}_{t+1}, \mathbf{\hat{X}}_{t+1}, \hat{\textbf{Y}}_{t+1}, P, Z, r_c, \texttt{local\_minimum}$\\

$V \gets$ Image-frame depth of $\mathbf{X}_{t+1}$ is less than corresponding value in $Z$\\

$C \gets d(\mathbf{X}_{t+1}, P) < r_c$\\
$\textbf{Y}_{t+1}[V \land \neg C] = 1$ \\
$\textbf{Y}_{t+1}[V \land C] = 0$\\
$I \gets \neg(V \land \neg C) \land (\hat{\textbf{Y}}_{t+1} < 0)$ \\
$\mathbf{X}_{t+1} = \mathbf{X}_{t+1}[(V \land C) \lor I \lor \texttt{local\_minimum}]$\\
$\mathbf{Y}_{t+1} = \mathbf{Y}_{t+1}[(V \land C) \lor I \lor \texttt{local\_minimum}]$\\
$\mathbf{\hat{X}}_{t+1} = \mathbf{\hat{X}}_{t+1}[I]$\\
$\hat{\textbf{Y}}_{t+1} = \hat{\textbf{Y}}_{t+1}[I]$
\caption{\texttt{pre\_process\_data}}
\label{alg:clean}
\end{algorithm}

\vspace{-.2cm}
\subsection{Adding Generated Data to GPIS}\label{sec:add}
While we calculate labels for the $2n$ points in $\mathbf{X}_{t+1}$ and $\mathbf{\hat{X}}_{t+1}$ at each timestep, we do not necessarily add all $2n$ points to the GPIS.
We choose which data points to add based on the progress being made by the controller and the semantics of the generated labels.

\subsubsection{Local Minima of Controller} \label{sec:local_min}
Due to the finite horizon of MPC, it is possible for the controller to enter a local minimum of the cost function from which it cannot make progress toward the goal.
As we will discuss later, we use the uncertainty of the GPIS as an exploration term in our MPC cost function to address this.
Adding data to the GPIS when the controller enters a local minimum changes the cost landscape by reducing the GPIS uncertainty at those points.
The change in the cost landscape can alleviate the local minimum, enabling further progress.
When a local minimum is detected, we add all components in $\mathbf{X}_{t+1}$ and their labels to the GPIS.
Note that this cannot add any interior points, as we do not add points in $\mathbf{\hat{X}}_{t+1}$.
To determine if we have entered a local minimum, we periodically check the average distance per timestep traveled by the manipulated object (line 8 in Alg. \ref{alg:cap}).
If this distance is below a threshold $d_{\mathrm{min}}$, we consider the last state to be at a local minimum.
We consider these data points ($\mathbf{X}_{t+1}$, $\mathbf{Y}_{t+1}$) separately from data points added to the GPIS when contact is inferred.
We define masks $M$ and $\bar{M}$ over all points in $D$ and $\bar{D}$ respectively, where $M^i=1$ if the $i$th data point corresponds to a detected local minimum and 0 otherwise.
This mask is used in our contact refinement in Section \ref{sec:contact_refine}.

\subsubsection{Visual Pre-processing}
\begin{figure}[tbp]
\includesvg[inkscapelatex=false, width=\linewidth]{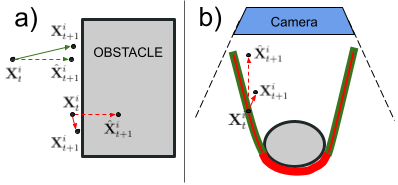}
\vspace{-.8cm}
\caption{\textbf{a)} The green transition has a small discrepancy between the predicted and actual next state, resulting in no contact detection. The red transition has a higher discrepancy, resulting in contact detection. \textbf{b)} When the cable is pulled upwards, the red transition would result in contact detection but does not due to visual pre-processing. Points that are updated by vision are indicated by the red line surrounded by green.}
\label{fig:label_gen}
\end{figure}

Only using dynamics and state estimates to generate labels can lead to false positive contacts along the full length of an object like a cable even if only a portion of the cable is in contact, as shown in Fig.~\ref{fig:label_gen}b.

To address this, we use visual input to clean the labels, as shown in Alg. \ref{alg:clean}.
Specifically, we determine if components are visible and, if so, whether they are in contact with the environment. We determine visibility by projecting component coordinates into image coordinates.
Given the intrinsic and extrinsic camera parameters, we can recover pixel coordinates $(u, v)$ for each component as well as a depth $z$ corresponding to the current state.
Letting $Z(u,v)$ be the depth value stored in the depth image $Z$ at $(u,v)$, a state component is visible if $z < Z(u,v)$.
Visible components are given a label of 1.

Components of $\mathbf{X}_{t+1}$ that are within a distance $r_c$ of a point in $P$ are considered as visibly in-contact with a label of 0.
$r_c$ is a parameter whose value is informed by the resolution of the point cloud and the geometry of the grasped object.

We add data to $D$ and $\bar{D}$ if a combination of critera are met: We add points in $\mathbf{\hat{X}}_{t+1}$ and their labels if they are non-visible interior points (lines, 6, 9 in Alg. \ref{alg:clean}) and we add points in $\mathbf{X}_{t+1}$ and their labels if they are visibly in-contact or correspond to interior points in $\mathbf{\hat{X}}_{t+1}$ (line 7 in Alg. \ref{alg:clean}).
We do not add the dynamics predictions of visible components to $D$ and $\bar{D}$ to avoid incorrectly adding interior points to the data sets.

\vspace{-.1cm}
\subsection{Visual Post-processing}
Even with pre-processed data, the specific values of kernel parameters and the nature of Gaussian process interpolation can lead to predictions of occupancy in known free space.
Along with using visual input to pre-process the labels in $D$ and $\bar{D}$, we also use the visual data to post-process the GPIS output.
When making a prediction, we check if the input point is visibly in free space. If so, we override the mean of the prediction and treat the point as being in free space.

As opposed to explicitly adding known free space points to the GPIS, this filtering step allows us to fit a useful surface while having an uncertainty landscape that only depends on the states visited during task execution.
As the Gaussian Process produces smooth uncertainty quantification, data corresponding to visible points can incorrectly reduce the uncertainty in non-visible areas.
This could be addressed through an additional input dimension corresponding to the visibility of the point, but we choose to implement the post-processing over the kernel engineering that would be required to include the semantic information as GPIS input.

\vspace{-.2cm}
\subsection{Contact Refinement}\label{sec:contact_refine}
A key contribution of COGIS is its ability to refine the estimate of the object using task-specific constraints. Due to noise in state estimation or dynamics as well as heuristics used in computing interior points, it is possible to fit models of the environment that do not satisfy the desired constraints.
For example, a surface may be fit that results in significant penetration of the state estimate into the surface, causing issues with planning methods.
In another case, there may be no paths to $G$ that avoid collision with $\hat{\mathcal{E}}$. 
As $\hat{\mathcal{E}}$ is parameterized by $\bar{D}$, we can improve constraint satisfaction by removing data from $\bar{D}$ that leads to constraint violation.
\begin{algorithm}[t]
Given $D$, $\bar{D}$, $\phi$, $T_{\texttt{CMA}}$, $N$, $h_{\text{all}}$\\
$D = D[M = 0]$ \\
$\bar{D} = \bar{D}[\bar{M} = 0]$ \\
$M = M[M = 0]$\\
$\bar{M} = \bar{M}[\bar{M} = 0]$\\
$\omega^* \gets \mathbf{1}^{|\bar{D}|}$ \\ 
$\phi^* = -1$ \\
$\texttt{CMA} \gets \mathrm{CMAwM}(\mathcal{N}(.5, .25))$\\
Calculate $c$ using \eqref{c-def}\\
\For{$T_{\texttt{CMA}}$ steps} { 
    $\Omega \gets \mathrm{Sample}(\texttt{CMA}, N)$\\
    Evaluate samples using $\phi$\\
    \For{$\omega_i \in \Omega$} {
        \If{
        $c^\top \omega_i > \phi^*$ \normalfont{and} $h_{\mathrm{all}}(\bar{D}, \omega_i)$
        } {
            $\omega^* = \omega_i$ \\
            $\phi^* = c^\top \omega_i$
        }
    }
    Update \texttt{CMA} with $\Omega$
}
Remove points from $\bar{D}$ where $\omega^* = 0$\\
\caption{\texttt{refine\_contacts}}
\label{alg:cma}
\end{algorithm}
We implement an integer optimization problem in \eqref{omega-opt} to refine $\hat{\mathcal{E}}$.

\vspace{-.2cm}
\begin{equation}\label{omega-opt}
\begin{aligned}
\omega^* = \argmax_{\omega} \quad & c^\top \omega\\
\textrm{s.t.} \quad & h_{\mathrm{all}}(\bar{D}, \omega, \dots) \\
  &\omega\in \{0, 1\}^{|\bar{D}|}    \\
\end{aligned}
\end{equation}
\begin{equation}\label{c-def}
\begin{aligned}
    c = \sigma(\sum_{i \in D} K(\bar{D}, D_i))
\end{aligned}
\end{equation}
\vspace{-.3cm}

\noindent $\omega \in \{0, 1\}^{|\bar{D}|}$ is a binary vector with an entry corresponding to each data point in $\bar{D}$, $c \in \mathbb{R}^{|\bar{D}|}$ is a weight for the optimization, described below, $K$ is the kernel function of the GPIS, and $\sigma$ is the softmax function.
A value of 0 corresponds to the data point being removed 
from $\bar{D}$. The ``$\dots$'' refer to auxiliary arguments that may be needed for computing various constraints.
This approach minimizes deviation from the current estimate by removing a minimal number of data points from $\bar{D}$ while ensuring constraint satisfaction.

While we seek to maximize the number of data points kept in $\bar{D}$, we bias the optimization toward keeping data points that have a higher density in $D$.
This reflects our assumption that points that are repeatedly encountered are less likely to be spurious.
By keeping a ``memory'' of all collected non-local minima data points, even ones that have been previously removed from $\bar{D}$, we can recover from incorrectly removing data points from $\bar{D}$ due to the distribution of $\bar{D}$ at the time of the previous optimization.
Through the use of this memory, we are able to reduce the search space of our optimization by searching over only the data points in $\bar{D}$ while still informing the optimization with all collected data.

$c$ is defined in \eqref{c-def} and biases the optimization toward keeping points in $\bar{D}$ that are more similar to points in $D$, with similarity calculated using $K$.
We use $\sigma$ to ensure a consistent scale for $c$ regardless of the size of $D$ or $\bar{D}$.

We solve this optimization problem using CMA-ES with Margin (CMAwM) \cite{cma}, as shown in Alg. \ref{alg:cma}.
CMAwM is a particle-based optimization method that handles constraints by including them in its objective function.
We construct an objective function $\phi$ in \eqref{cma-objective} minimized by CMAwM that includes the objective function from \eqref{omega-opt} as the first term and a constraint violation penalty as the second term:]

\vspace{-.5cm}
\begin{equation}\label{cma-objective}
    \phi(\bar{D}, \omega, \dots) = -c^\top \omega + 10(1-h_{\mathrm{all}}(\bar{D}, \omega, \dots))
\end{equation}
\vspace{-.6cm}

As CMAwM does not require gradients to optimize $\phi$, we can use constraints that may not be convex or differentiable with respect to $\omega$.

After the optimization has run, states that may have been local minima given the previous $\hat{\mathcal{E}}$ may no longer be local minima.
Keeping the previously detected local minima in $\bar{D}$ may prevent the controller from exploring states needed to collect contact data as they would have low variance under the GPIS.
To address this, we remove points from $D$ and $\bar{D}$ corresponding to local minima using the masks $M$ and $\bar{M}$ when the optimization is triggered (lines 2-5, Alg. \ref{alg:cma}).

\vspace{-.4cm}
\subsection{Controller}\label{sec:control}
Our MPC cost function is $J(\tau) = J_g(\tau) + \alpha J_u(\tau) + C J_c(\tau) + \beta J_e(\tau)$ where the different terms are: a goal directed cost $J_g$, an action regularization cost $J_u$, a collision cost $J_c$, and an exploration cost $J_e$.
$\alpha, \beta, C \in \mathbb{R}$ are coefficients used to weigh the different costs.

\subsubsection{Goal Cost}
The goal cost in \eqref{J_g} drives the controller toward the goal state. $d_\mathbf{x}(G_i, \textbf{X}_t^i)=0$ if there is no goal defined for component $i$.
If only this distance is used in the cost, it is possible for other cost terms to overwhelm the goal cost near the goal, preventing the trajectory from converging successfully.
We add a cost term that helps the controller converge to the goal by creating a deeper basin in the cost function near the goal that the controller can exploit.
We use an indicator $\mathds{1}_{g_t}$, which is 1 when the goal distance is less than $r_g$ for all state components with defined goals.
$r_g$ is the distance from the goal that would indicate task success.
We weight this term by a parameter $\eta \in \mathbb{R}$.

\begin{equation}
    \label{J_g}
    J_g(\tau)= \sum\limits_{t=1}\limits^{T}\left(- \eta \cdot \mathds{1}_{g_t} + \sum\limits_{i\in G} d_\mathbf{x}(G_i, \textbf{X}_t^i)\right)
\end{equation}

\subsubsection{Action Regularization Cost}
$J_u(\tau) = \sum\limits_{t=0}\limits^{T-1} ||u_t||_2$ penalizes large actions to encourage smooth motion.

\subsubsection{Collision Cost}
We define a collision cost in \eqref{J_c} that penalizes transitions that would cause collision with $\hat{\mathcal{E}}$.
This is done by calculating the posterior prediction of the GPIS for the states along a rollout.

\vspace{-.4cm}
\begin{equation}
    \label{J_c}
    J_c(\tau) = \sum\limits_{t=1}\limits^{T}\sum\limits_{i=1}\limits^n \mathds{1}_{\hat{\mathcal{E}}(\mathbf{x}^i_{t}) \leq 0}
\end{equation}

\subsubsection{Exploration Cost}
As in prior work, we use the uncertainty to gain information about the obstacles by exploring new regions of state space.
We also find this form of exploration useful to help escape from what would otherwise be local minima of the controller.
The exploration cost $\label{J_e}
J_e(\tau) = -\sum\limits_{t=1}\limits^{T}{\sigma^s_t}^2$ uses the variance of the GPIS at a state $\mathbf{X}$, where ${\sigma_t}^2 \in \mathbb{R}^n$ is the variance of the normal distribution predicted by the GPIS for the $n$ components.

\begin{figure}[tbp]
\centerline{\includegraphics[width=\linewidth]{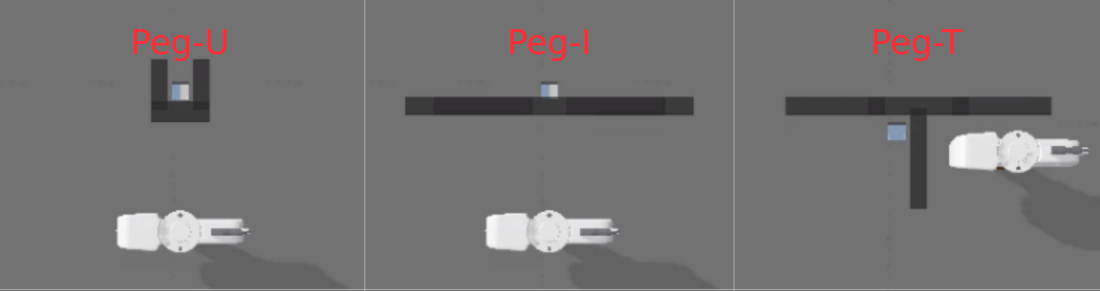}}
\vspace{-.2cm}
\caption{The peg-in-hole environments. The end-effector is grasping a peg, which the robot navigates to the hole. Environments and figure are from \cite{TAMPC}.}
\label{fig:peg_envs}
\end{figure}

\vspace{-.4cm}
\section{Results}
We evaluate our method on peg-in-hole and deformable object manipulation tasks, demonstrating the method's utility for manipulating objects with varying state dimensions and task requirements.
We use model predictive path integral control (MPPI) \cite{MPPI} for MPC.
We use a Matern kernel for the GPIS with $\nu=1.5$.
Other parameter values are provided in Table II.
We use \cite{gardner2018gpytorch} to implement the Gaussian process.

\subsection{Constraint Optimization Implementation}
For the following experiments, we provide definitions of constraints we enforce on the surface.
These constraints are violated in the presence of spurious interior points.
The presence of exterior points does not increase the likelihood that these constraints are violated.
Therefore, in our CMAwM optimization, we only optimize $\omega$ for the interior points in $\bar{D}$, and pre-fix the values in $\omega$ corresponding to exterior points to 1.
This is consistent with maximizing \eqref{omega-opt} and allows us to reduce the size of our search space at runtime.

\subsection{Peg-in-Hole}

\begin{figure}[tbp]
\centerline{\includesvg[inkscapelatex=false, width=\linewidth]{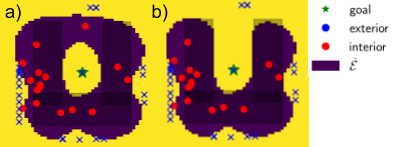}}
\vspace{-.2cm}
\caption{Predicted obstacles for the Peg-U task. \textbf{a)} A constraint violation;  the goal is encompassed by the obstacle, violating the constraint. \textbf{b)} Optimized surface; CMAwM removes the red interior point above the goal, satisfying the constraint.}
\label{fig:peg_before_after}
\end{figure}

We use the peg-in-hole tasks defined in \cite{TAMPC} and shown in Fig. \ref{fig:peg_envs}.
In these tasks, an end-effector simulated in PyBullet \cite{Pybullet} navigates a peg to a goal hole.
We assume the goal location is known but assume no prior knowledge of obstacles, necessitating adaptation.
Due to heuristic placement of interior points, the narrow opening in the Peg-U task can induce surfaces that block the path to the goal, as shown in Fig.~\ref{fig:peg_before_after}.
A success is defined as placing the peg within 2cm of the hole within 750 control steps.

The state is $(x,y, R_x, R_y)$, where $(x,y)$ represents the $\mathbb{R}^2$ end-effector position and $(R_x, R_y)$ are reaction forces.
The control signal is $(\Delta x, \Delta y)$.
We execute 1 step of a planned trajectory, replanning at each timestep.
We only consider the $\mathbb{R}^2$ position to fit the surface.
For these tasks, $n=1$.

\subsubsection{Constraint}

For these tasks, we enforce a constraint using connected components to guarantee the existence of a collision-free path between the goal and the tracked point at the center of the peg.
We calculate connected components of a binary image of the scene based on the GPIS predicted semantics using \cite{cc_torch}.
If the current state and the goal state are in different components, the constraint is violated.

\subsubsection{Analysis}
We compare our method to TAMPC and use their pre-trained dynamics model for $f$, which is learned without the presence of obstacles.
We do not use visual input for our method to provide a comparison to TAMPC.
We also evaluate two ablations: one with \texttt{refine\_contacts} ablated and one with no local minima detection.

We use the MPPI parameters in \cite{TAMPC} and tune the parameters for our method with Bayesian optimization.

As shown in Table I, we achieve similar to higher success rates than TAMPC and the ablations on these tasks over 30 trials.
We believe this is due to the learned surface providing a dense geometric model of the environment that is informed by topological constraints, allowing for collision checking while guaranteeing task feasibility.
We also show the utility of the contact refinement optimization for the Peg-U task through a higher success rate when compared to the ablation.

We show comparable results between the ablation and full method for the Peg-T task.
This is expected as the geometry for the Peg-T and Peg-I tasks is less likely to induce violations of the connected components constraint.
There is some increase in success rate for the Peg-I task, caused by the constraint sometimes being violated when there is penetration of the peg state into the surface estimate.
As we remove data points corresponding to local minima from $\bar{D}$ before running CMAwM, this can result in a ``reset'' of the exploration cost, making it more likely for the controller to succeed.
The learned surface also approximates the true obstacle geometry, as seen in Fig.~\ref{fig:peg_before_after}.

\vspace{-.4cm}
\subsection{Simulated Cable Manipulation}

\begin{figure}[tbp]
\centerline{\includegraphics[width=\linewidth]{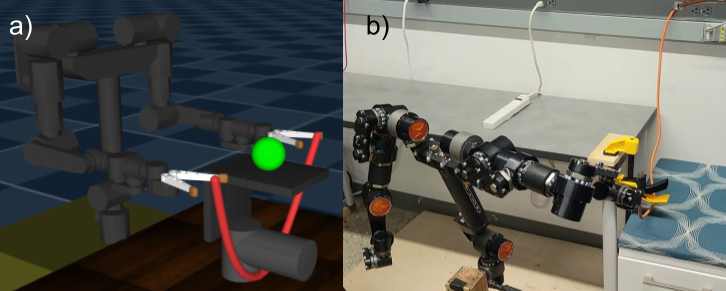}}
\vspace{-.2cm}
\caption{Deformable object tasks. \textbf{a)} Simulated cable manipulation task. The robot navigates the center of the cable to the green goal region. The camera is located above and behind the robot, causing the obstacle geometry to induce a high degree of occlusion. \textbf{b)} Real world cable manipulation task. The robot starts grasping the cable under the clamp and navigates it to the power strip.}
\label{fig:hook_visibility}
\end{figure}

In this task, a two-armed, 16-dof robot removes a cable from under a hook. The hook has a barrier that occludes part of the obstacle, as shown in Fig. \ref{fig:hook_visibility}.
The high degree of occlusion motivates estimating the obstacle geometry and the contact refinement.
A success is placing the center of the cable in a 4cm radius sphere over the hook within 200 steps.

We use a MuJoCo \cite{todorov2012mujoco} simulation to model $f$.
We include the observed environment in our dynamics by constructing a mesh from $P$ and using it in MuJoCo.
We represent the cable as 25 articulated links and track the $\mathbb{R}^3$ position for each link.
These positions comprise the state components.
The control is $[\Delta p_l, \Delta p_r] \in \mathbb{R}^6$, where $\Delta p_l$ is the change in the left gripper's position and $\Delta p_r$ is the change in the right gripper's position.

We use CDCPD2 \cite{CDCPD} in combination with the state from the simulation of the partial environment to estimate the cable state. CDCPD2 includes regularization terms that promote smoothness of the estimate and prevent large deviations in the estimate between timesteps, leading to reasonable estimates for occluded portions of the cable.

\subsubsection{Constraint}
We constrain the surface to prevent penetration of the cable state estimate by calculating GPIS predictions for each state component.
We find it useful to be conservative by calculating a lower bound on the GPIS prediction using the uncertainty.
Specifically, if $\mu$ and $\sigma$ are the mean and variance of the Gaussian predicted by the GPIS, then we check if $\mu + \Phi^{-1}(\zeta)\sigma \leq 0$, where $\Phi$ is the CDF of the standard normal distribution and $\zeta \in (0, 1)$. $\zeta=.4$ for this task.

\subsubsection{Analysis}
We compare our method to TAMPC and a baseline which directly uses the partial visual information without any online adaptation.
We also run ablations, individually ablating the local minima data addition, the visual pre-processing, visual post-processing, and the contact refinement step.
For the TAMPC state distance function, we consider the $\mathbb{R}^3$ position of the center of the cable to provide a more useful distance than a distance in the full $\mathbb{R}^{75}$ state, which we found to be ineffective.
We do not train a residual dynamics model online for TAMPC as the online data is insufficient for training a useful model for the high-dimensional state-action space.

Our results in Table I show that our method achieves higher success rates than the baselines and contact refinement ablation over 30 trials.
We also show that the contact refinement provides the largest increase in performance of the various design choices.
The ablated method can fail if spurious interior points close off the gap between the hook and the table.
Our contact refinement algorithm can remove the spurious data points, improving task success.
We believe TAMPC's trap representation provides a sparser signal to the controller and struggles to cover the space of possible local minima induced by the hook.
Our method reasons about contacts along the length of the cable independently, enabling us to learn a richer model of the environment that improves task performance.
The non-adaptive baseline cannot reason about the occluded part of the obstacle, leading it to collide repeatedly with the obstacle.

\begin{table}[tbp]
\label{tab:results}
\begin{center}
\begin{tabular}{|c|c|c|c|}
\hline
\multirow{2}{*}{Environment} & \multirow{2}{*}{Method}& \multirow{2}{*}{Success} & Control Steps \\
& & & (Given Success)\\
\hline
\multirow{4}{*}{Peg-U} & COGIS (Ours)& 26/30 & $296.3 \pm 84.4$\\
& COGIS-No local minima& 12/30 & $\mathbf{ 129.17 \pm 33.4}$\\
& COGIS-No refinement & 11/30 & $135.7 \pm 62.9$\\
& TAMPC&  \textbf{27/30} & $248.3 \pm 57.4$\\
\hline
\multirow{4}{*}{Peg-I} & COGIS (Ours)& \textbf{27/30} & $328.7 \pm 56.9$\\
& COGIS-No local minima& 3/30 & $ 499.7 \pm 248.9$\\
& COGIS-No refinement& 25/30 & $316.2 \pm 57.3$\\
& TAMPC& 23/30 & $\mathbf{274.4 \pm 45.6}$\\
\hline
\multirow{4}{*}{Peg-T} & COGIS (Ours)& \textbf{30/30} & $123.3 \pm 35.1$\\
& COGIS-No local minima& 25/30 & $ 234.2 \pm 61.7$\\
& COGIS-No refinement& \textbf{30/30} & $\mathbf{107.7 \pm 26.9}$\\
& TAMPC& 25/30 & $160.3 \pm 25.0$\\
\hline
\multirow{7}{*}{Sim. Cable} 
& COGIS (Ours)& 23/30& $114.4 \pm 21.8$\\
& COGIS-No local minima& \textbf{24/30} & $ 122.8\pm 13.8$\\
& COGIS-No vis. pre-process& 22/30 & 135.7 $ \pm 15.1$\\
& COGIS-No vis. post-process& 21/30 & $121.4 \pm 13.6$\\
& COGIS-No refinement &  17/30& $\mathbf{111.5 \pm 22.7}$\\
& TAMPC& 1/30 & $161 \pm 0$\\
& Non-adaptive & 1/30 & $174 \pm 0$\\
\hline
\multirow{3}{*}{Real Cable} 
& COGIS (Ours)& \textbf{10/10} & $\mathbf{74.1 \pm 5.8}$\\
& COGIS-No refinement &  9/10 & $88.4 \pm 11.4$\\
& TAMPC&  1/10 &  $132 \pm 0$\\
\hline
\end{tabular}
\end{center}
\caption{Success rates, 95\% confidence intervals for control steps. Control step statistics are calculated for successful trials.}
\vspace{-.3cm}
\end{table}

\vspace{-.3cm}
\subsection{Real Cable Manipulation}
In this task, the robot navigates an extension cord around an occluded obstacle to a goal power strip.
Part of a clamp underneath the table is occluded, as shown in Fig. \ref{fig:hook_visibility}, requiring online adaptation to successfully complete the task.
The higher amount of state estimation noise in the real world along with the gap between the MuJoCo dynamics used in MPPI and the true cable dynamics can lead to artifacts in the estimated surface which motivate contact refinement.

We again use a MuJoCo model of the partially observed environment for $f$ with the same cable state.
For this task, one gripper is used as one end of the cable is fixed to the wall.
We use the same constraint as used in the simulated cable experiment, with $\zeta=.45$.

\subsubsection{Analysis}
We find that both the ablation and full method are able to consistently solve the task over 10 trials, but we observe qualitative differences in the estimated surfaces and quantitative differences in the episode lengths.
As can be seen in Fig.~\ref{fig:grabber}, without the additional optimization, error in state estimation or dynamics can create artifacts in the surface estimate.
These artifacts can lead to longer trajectories, as seen by the greater number of actions taken by the ablation.

Our method outperforms TAMPC, which struggles due to the high dimensional nature of the problem and the potentially long recovery horizon.

\vspace{-.2cm}
\section{Discussion and Conclusion}
The main limitation of our method is its assumption that dynamics error is caused by contact.
This can lead to placing surfaces in regions where there is no unobserved obstacle due to improper modeling of the physical system.
This improper modeling could be due to incorrect estimations of system properties, for example the stiffness or other physical parameters of a cable, or due to the use of learned dynamics or low-fidelity simulators.
Our contact refinement step is capable of mitigating this through imposing constraints that can remove artifacts generated due to non-contact based dynamics error.
However, it is possible that large deviations from nominal dynamics, for example due to a large gap between the simulator and reality, could generate more spurious obstacles than the refinement is able to compensate for.

As the constraints are task-informed, it is possible that a surface fit for one task may result in poor performance if used as is in another task.
However, COGIS should be able to generate surfaces online given meaningful constraints for a new task.
While we cannot guarantee that we will only detect true contacts with the environment, we show in our results in Table I that the inclusion of the contact refinement leads to higher task success rates.

We presented Constraint Obeying Gaussian Implicit Surfaces (COGIS), a method for modeling \textit{a priori} unknown obstacles while ensuring these models satisfy desired constraints. Through this we enable rapid adaptation of manipulation to partially observable environments.
We achieve higher success rates than baselines and ablations across multiple tasks, including high-dimensional deformable object manipulation tasks.
Our method leverages a novel fusion of visual and inferred contact information to model obstacles using a Gaussian process implicit surface along with a novel contact refinement step, enabling data-efficient obstacle modeling for use in MPC. 

\vspace{-.3cm}
\section{Acknowledgments}
The authors would like to thank Dylan Colli for his help with CDCPD2.
\vspace{-.5cm}

\bibliographystyle{IEEEtran}
\bibliography{ref.bib}

\begin{table}[tp]
\label{mppi_param}
\begin{center}
\begin{tabular}{|l|c|c|c|}
\hline
& Peg-in-Hole & Sim. Cable & Real Cable\\
\hline
$\lambda$ MPPI temperature& .01 & .167 & .5\\
\hline
$K$ MPPI samples& 500 & 72 & 55\\
\hline
$T$ MPPI horizon& 10, 15, 20 & 8 & 12\\
\hline
$\Sigma$ MPPI noise& $\mathbf{diag}[.2_{\times2}]$ & $\mathbf{diag}[.004_{\times6}]$ & $\mathbf{diag}[.003_{\times6}]$\\
\hline
$\alpha$& .590 & .627 & 10\\
\hline
$\beta$& .996 & .995 & .4\\
\hline
$\eta$& 11.03 & 100 & 1000\\
\hline
$C$ & 15.88 & 10000 & 10000 \\
\hline
$d_{min}$ & .01 & .01 & .0025 \\
\hline
$T_m$& 5 & 3 & 1\\
\hline
$T_e$& - & 3 & 3\\
\hline
$T_{\mathrm{fit}}$ & 3 & 2 & 2 \\
\hline
$r_g$& .02 & .04 & .1\\
\hline
$r_c$& - & .01 & .01\\
\hline
$T_{\texttt{CMA}}$ & 25 & 25 & 25\\
\hline
$N$ & 20 & 50& 50\\
\hline
\end{tabular}
\label{tab:gpis_hyper}
\end{center}
\caption{Parameters}
\end{table}
\end{document}